\pdfoutput=1
\documentclass[11pt]{article}

\usepackage[]{acl}

\usepackage{times}
\usepackage{latexsym}

\usepackage{pgfplots}
\usepackage{scalefnt}
\usepackage{amsfonts}
\usepackage{amsmath,bm}
\usepackage{mathrsfs}
\usepackage{multirow}
\usepackage{verbatim}
\usepackage{amssymb}
\usepackage{booktabs}
\usepackage{subfigure}

\usepackage[T1]{fontenc}
\usepackage[utf8]{inputenc}

\usepackage{microtype}

\title{CTRLEval: An Unsupervised Reference-Free Metric for Evaluating Controlled Text Generation}

\author{Pei Ke$^1$, Hao Zhou$^2$, Yankai Lin$^2$, Peng Li$^3$\thanks{\quad Part of the work was done while Peng Li was working at Tencent.}, Jie Zhou$^2$, Xiaoyan Zhu$^1$, Minlie Huang$^1$\thanks{\quad Corresponding author} \\
\small $^1$The CoAI group, DCST, Institute for Artificial Intelligence, State Key Lab of Intelligent Technology and Systems, \\
\small Beijing National Research Center for Information Science and Technology, Tsinghua University, Beijing 100084, China \\
\small $^2$Pattern Recognition Center, WeChat AI, Tencent Inc., China \\
\small $^3$Institute for AI Industry Research (AIR), Tsinghua University, China \\
\tt\small kepei1106@outlook.com, \{tuxzhou,yankailin,withtomzhou\}@tencent.com\\ 
\tt\small lipeng@air.tsinghua.edu.cn, \{zxy-dcs,aihuang\}@tsinghua.edu.cn \\
  }

\begin{document}
\maketitle
\begin{abstract}

%Recent works on controlled text generation directly utilize pre-trained models to generate texts towards a desired attribute without fine-tuning or re-training them on the downstream datasets with attribute annotation, which is challenging for evaluation. Without human references, existing metrics which correlate well with human ratings are mostly trained on the task-specific datasets, which may cause over-fitting and degrade the generalization ability. 

%Existing works on controlled text generation mostly adopt the automatic metrics trained on task-specific datasets to evaluate different aspects of generated texts, which may cause over-fitting and degrade the generalization ability.
%Supervised metrics may overfit the training data and degrade the generalization ability of evaluating unseen samples, whereas unsupervised ones can only provide a non-task-specific evaluation result, which correlates weakly with human judgments.
%Supervised metrics may overfit the training data and degrade the generalization ability of evaluating unseen samples, whereas unsupervised ones can only provide a non-task-specific evaluation result, which correlates weakly with human judgments.
Existing reference-free metrics have obvious limitations for evaluating controlled text generation models. Unsupervised metrics can only provide a task-agnostic evaluation result which correlates weakly with human judgments, whereas supervised ones may overfit task-specific data with poor generalization ability to other datasets. In this paper, we propose an unsupervised reference-free metric called \textit{CTRLEval}, which evaluates controlled text generation from different aspects by formulating each aspect into multiple text infilling tasks. On top of these tasks, the metric assembles the generation probabilities from a pre-trained language model without any model training. Experimental results show that our metric has higher correlations with human judgments than other baselines, while obtaining better generalization of evaluating generated texts from different models and with different qualities\footnote{The data and codes are available at \url{https://github.com/thu-coai/CTRLEval}.}.

%We argue that it's challenging to evaluate current methods

%the lack of evaluation metrics has increasingly hindered the development

\end{abstract}

% 1. assign specific aspects to unsupervised metrics
% 2. multiple evaluators

\section{Introduction}

Controlled text generation aims to generate texts under some control variables, including pre-specified content prefixes and attribute labels (such as sentiments and topics). Controlled text generation has been significantly advanced by large-scale pre-trained models with respect to generation quality and various control variables ~\cite{keskar2019ctrl,da2020pplm,yang2021fudge,liu2021dexpert,chan2021cocon}. 

Despite the great success of these generation models, it becomes critical to evaluate the quality of generated texts accurately.
%for existing evaluation metrics to distinguish the quality of generated texts.
%Compared with the rapid development of model design, automatic evaluation of controlled text generation models is still under-explored. 
Most of the existing studies adopt unsupervised and supervised metrics to measure the quality of generated texts under different combinations of control variables \cite{da2020pplm,chan2021cocon}. 
The evaluation is commonly conducted in a reference-free setting because it is challenging to collect sufficient high-quality references for each input of control variables in this open-ended text generation task \cite{da2020pplm}.
%Since the conditions are manually selected by humans, the evaluation is commonly conducted in a reference-free setting \cite{da2020pplm}.

However, both unsupervised and supervised metrics have shown limitations in the evaluation of controlled text generation: 1) Unsupervised metrics such as perplexity \cite{brown1992perplexity} can only provide task-agnostic evaluation regarding the overall quality of generated texts. However, controlled text generation tasks typically involve multiple evaluation aspects \cite{deng2021unified}, including the quality of generated texts themselves and the relationship between generated texts and control variables.
%\textit{coherence} within the generated text, \textit{consistency} between input and ouput, and \textit{faithfulness} of implementing the control variables. 
It is thus not surprising that existing unsupervised metrics without multi-aspect interpretability have low correlations with human judgments \cite{hashimoto2019unify}. 
2) Supervised metrics are commonly trained on the datasets of specific tasks to measure the corresponding aspects of generated texts (e.g., evaluating whether a generated text is accordant with the sentiment label) \cite{da2020pplm,chan2021cocon}. This may cause over-fitting to task-specific data and degrade the generalization ability of metrics \cite{garbacea2019judge}, thereby giving unstable evaluation of generated texts from different models 
or with different qualities \cite{guan2020union}.

To deal with the above issues, we propose an unsupervised reference-free metric called \textit{CTRLEval} for evaluating controlled text generation models. This metric performs evaluation from different aspects without any training on task-specific data. Specifically, we formulate the evaluation of each aspect into ``fill-in-the-blank'' tasks whose input and output patterns can be designed based on the definition of the aspect. Then, we utilize a pre-trained model whose pre-training task is text infilling (such as PEGASUS \cite{zhang2020pegasus}) as our base model, and fuse the generation probabilities from these ``fill-in-the-blank'' tasks as the evaluation result. 
%In our framework, each text infilling task is denoted as a \textit{pattern evaluator}, which means evaluation with different input and output patterns. 
To alleviate the potential bias caused by the task design \cite{zhao2021calibrate}, we devise multiple text infilling tasks for each aspect and use the weighted sum of all the results as the final score.
%The ensemble of multiple pattern evaluators for each aspect is expected to alleviate the potential bias brought by the pattern design \cite{zhao2021calibrate}. 
In this paper, we consider three aspects which are commonly used to measure the performance of controlled text generation models,
%and devise multiple text infilling tasks for each of them, 
including coherence \cite{yuan2021bartscore}, consistency \cite{rashkin2020plotmachine}, and attribute relevance \cite{da2020pplm}. These evaluation aspects cover both the quality of generated texts and the relationship between generated texts and different control variables, which can provide a comprehensive evaluation result for controlled text generation.
Experimental results show that our metric can maintain the generalization ability and achieve stable performance faced with
%on the generated texts 
model drift and quality drift.
Our main contributions are as follows:
\begin{itemize}
    \item We propose an unsupervised reference-free metric called CTRLEval for evaluating controlled text generation. This metric formulates three evaluation aspects (i.e., coherence, consistency, and attribute relevance) into multiple text infilling tasks, and utilizes the ensemble of generation probabilities from a pre-trained language model as the evaluation results.
    
    \item We conduct experiments on two benchmark tasks including sentiment-controlled and topic-controlled text generation based on our collected evaluation set. Experimental results show that our proposed metric has higher correlations with human judgments, while obtaining better generalization of evaluating generated texts from different models and with different qualities.
\end{itemize}

%Despite the trend of controlled text generation based on pre-trained models without fine-tuning, we argue that the lack of effective evaluation metrics has increasingly hindered the development of this research direction. There are two major challenges for evaluating controlled text generation: 1) \textbf{Reference-free}: Since most of the existing work directly makes the pre-trained models generate texts conditioned on the given attribute label and content prefix, there is no human reference for evaluation. Thus, the metrics based on lexicon overlap (such as BLEU \cite{papi2002bleu}) or semantic similarity (such as BERTScore \cite{zhang2020bertscore}) between generated hypothesis and human reference are not proper to evaluate controlled text generation. 2) \textbf{Generalization}: Existing reference-free metrics which correlate well with human evaluation results commonly need to be trained on the dataset to fit the human score \cite{sellam2020bleurt} or distinguish the human-written texts from manually created negative samples \cite{guan2020union}. This may lead to over-fitting to specific data because recent controlled text generation models don't need to fine-tune on the downstream corpora.

\section{Related Work}

\subsection{Controlled Text Generation}

Early studies on controlled text generation adopt attribute label embeddings \cite{ficler2017control,zhou2018ecm} or latent variables \cite{hu2017control,ke2018senfunc,zhou2018moji} to learn the complex relationship between control variables and generated texts. With the development of large-scale generative pre-trained models, 
%CTRL \cite{keskar2019ctrl} trains a large-scale language model with 1.6B parameters to generate texts conditioned on a variety of control codes. Since 
it is costly to re-train or fine-tune pre-trained models on the corpora with attribute annotations \cite{keskar2019ctrl}. 
Recent works resort to decoding-time methods and directly make pre-trained models generate texts towards desired attributes during inference, including PPLM \cite{da2020pplm}, GeDi \cite{krause2020gedi}, FUDGE \cite{yang2021fudge} and DEXPERTS \cite{liu2021dexpert}. 
%These works devise various attribute discriminators to modify the hidden states or generation probabilities of generators. 
%The evaluation protocol of these works is to make the models generate texts conditioned a content prefix and an attribute label, and then evaluate different aspects of generated texts in a reference-free setting \cite{da2020pplm}.
These works rely heavily on human evaluation because existing reference-free metrics including unsupervised and supervised ones are shown to have evident limitations for evaluating controlled text generation \cite{da2020pplm}.
%thereby guiding text generation towards pre-specified attributes.
%For instance, PPLM \cite{da2020pplm} updates the latent hidden states of the pre-trained models using the gradients from the attribute model and samples the next token from the updated generation probability distribution. GeDi \cite{krause2020gedi} uses class-conditioned language models as generative discriminators and efficiently computes the classification likelihoods over all the candidate next tokens using Bayes rules. FUDGE \cite{yang2021fudge} proposes a future discriminator to determine whether the attribute will be achieved in the future based on an incomplete sentence prefix and adopts this discriminator to guide language generation. DEXPERTS \cite{liu2021dexpert} combines an out-of-the-box pre-trained models with expert models and anti-expert models, which model texts with desirable and undesirable attributes, respectively.

%Another line of works like CoCon \cite{chan2021cocon} utilizes self-supervised learning approaches to train a block with fewer parameters which incorporates the content of attributes into the output of large-scale pre-trained models.

%Since existing works put more emphasis on the design of generation models, our work focuses on the evaluation of controlled text generation and proposes an unsupervised reference-free metric to evaluate the generation quality from the perspectives of coherence, consistency and controllability.

\begin{figure*}[!htp]
  \centering
  \includegraphics[width=1.0\linewidth]{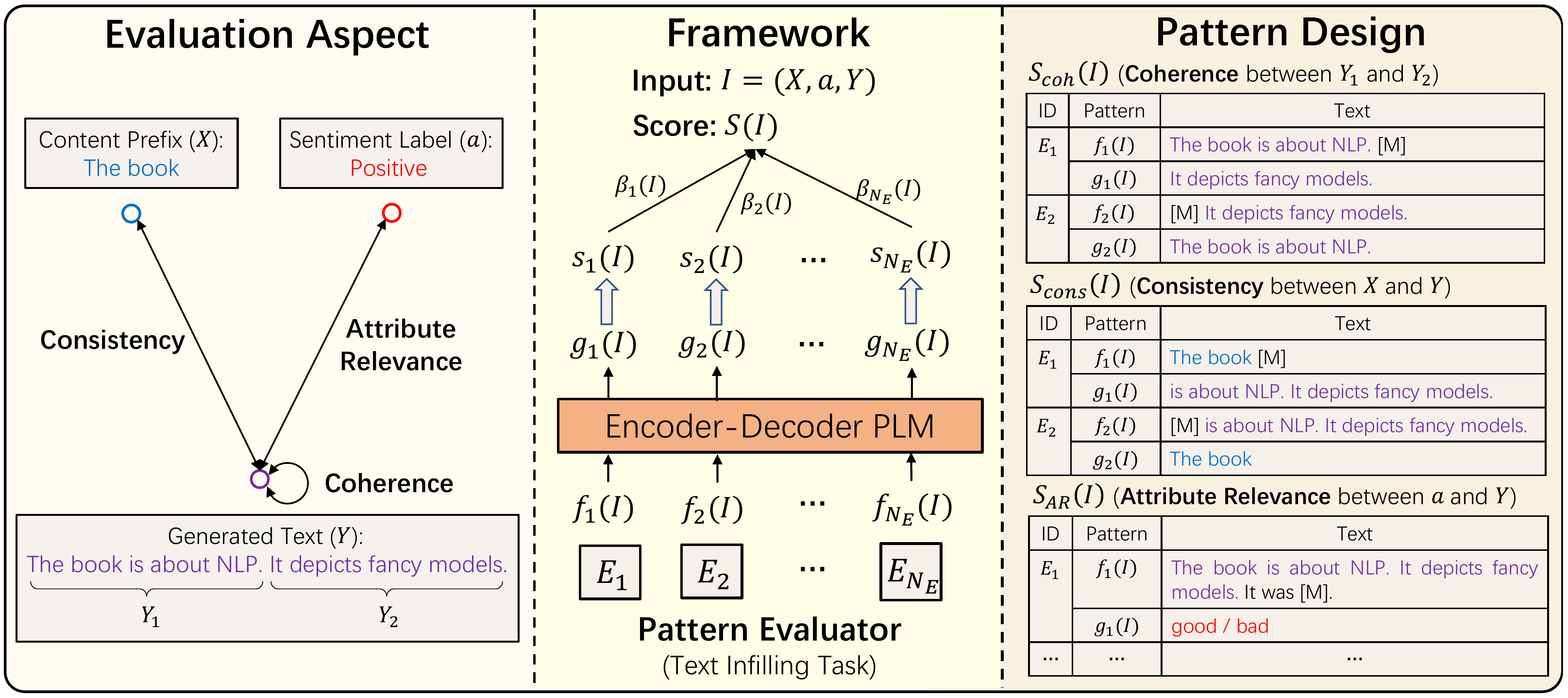}% 1\linewidth
  \caption{Overview of CTRLEval. \textbf{Left}: The three evaluation aspects measure the relationship among content prefixes, attribute labels, and generated texts. \textbf{Medium}: The evaluation result $S(I)$
  %the three aspects $S_{coh}(I)$ / $S_{cons}(I)$ / $S_{AR}(I)$ 
  is computed based on the ensemble of the scores from pattern evaluators $E_j(1\leq j\leq N_E)$. The score $s_j(I)$ of each pattern evaluator $E_j$ is obtained by the generation probability of the encoder-decoder pre-trained language model in the text infilling task, with the input of $f_j(I)$ and the output of $g_j(I)$. \textbf{Right}: The evaluation results for three aspects $S_{coh}(I)$ / $S_{cons}(I)$ / $S_{AR}(I)$ are acquired by the corresponding pattern evaluators, respectively.}
  %\caption{Overview of CTRLEval. \textbf{Left}: The three evaluation aspects measure the relationship among content prefixes, attribute labels and generated texts. \textbf{Right}: The evaluation results of the three aspects $S_{coh}(I)$ / $S_{cons}(I)$ / $S_{AR}(I)$ are computed based on the ensemble of the scores from corresponding pattern evaluators $E_j(1\leq j\leq N_E)$. The score $s_j(I)$ of each pattern evaluator $E_j$ is obtained by the generation probability of the encoder-decoder pre-trained language model in the text infilling task, with the input of $f_j(I)$ and the output of $g_j(I)$.}
  \label{fig:ctrlmodel}
\end{figure*}

\subsection{Evaluation Metric for Text Generation}

Automatic evaluation metrics are important for natural language generation tasks, which can be simply divided into referenced, reference-free (also known as unreferenced) and hybrid metrics: 1) Referenced metrics usually measure the relevance between generated texts and reference texts via lexicon overlap (such as BLEU \cite{papi2002bleu}, METEOR \cite{banerjee2005meteor} and ROUGE \cite{lin2004rouge}) or embedding similarity (such as MoverScore \cite{zhao2019moverscore}, BERTScore \cite{zhang2020bertscore} and MARS \cite{liu2021mars}).
%Since these referenced metrics poorly correlate with human ratings particularly in the open-ended text generation tasks, MARS  augments the reference by filling the masked tokens of reference texts via reinforcement learning.
2) Reference-free metrics directly evaluate the quality of generated texts without references. Since unsupervised metrics like perplexity \cite{brown1992perplexity} and distinct n-grams \cite{li2016diversity} can only provide a task-agnostic result which correlates weakly with human judgments \cite{hashimoto2019unify,tevet2021diversity}, most of the reference-free metrics resort to supervised models. Specifically, they are trained to fit human-annotated ratings / labels (such as discriminator scores \cite{shen2017style}) or distinguish human-written texts from negative samples (such as UNION \cite{guan2020union}). 3) Hybrid metrics contain both referenced and reference-free scores, such as RUBER \cite{tao2018ruber,sarik2019ruberbert}, BLEURT \cite{sellam2020bleurt} and BARTScore \cite{yuan2021bartscore}.

%Compared with existing works, our method falls into the category of unsupervised reference-free metrics. By formulating the evaluation criteria into the text infilling tasks, our metric can fully utilize the power of pre-trained models to achieve promising evaluation results with better generalization ability.

Compared with existing reference-free metrics which are unsupervised, our metric can support the evaluation of generated texts from different aspects via the full utilization of pre-trained models and the formulation of text infilling tasks, which fits the evaluation protocol of controlled text generation well. Also, in contrast with supervised reference-free metrics, our metric can avoid over-fitting task-specific data and maintain better generalization ability to evaluate generated texts from different models and with different qualities.

\section{Method}

\subsection{Task Definition and Method Overview}

Given the input $I=(X,a,Y)$ which consists of a content prefix $X$, an attribute label $a$, and a generated text $Y$, our goal is to acquire three evaluation results for coherence, consistency and attribute relevance, respectively.

As shown in Figure \ref{fig:ctrlmodel}, our main idea is to formulate each evaluation aspect into multiple text infilling tasks and utilize the ensemble of the scores from each task as the final evaluation results.
%generation probability of pre-trained models as the evaluation results.
%the ensemble of language model scores from each task as the evaluation result. 
We denote each text infilling task as a \textit{pattern evaluator}
%\footnote{This means evaluation with different input / output patterns.} in our paper.
, which means evaluation with different input and output patterns.
Inspired by the recent works on pattern-exploiting training \cite{schick2021pet,schick2020petforgen} and prompt tuning \cite{gu2021ppt}, we define each pattern evaluator as $E=(f,g)$, which consists of two pattern functions to build the input and output sequence of text infilling tasks, respectively.
The score of each pattern evaluator is acquired from the generation probability of the encoder-decoder pre-trained language model whose pre-training task is to generate the masked part from the remaining texts of the input.
%, such as PEGASUS \cite{zhang2020pegasus}.
%\begin{itemize}
    %\item Input Template Function $f(\cdot)$: This template function maps the input to a text sequence with mask tokens and prompts, which is used for the input of pre-trained models.
    %\item Output Template Function $g(\cdot)$: This template function maps the input to a text sequence which is used for the output of pre-trained models.
    %\item Verbalizer $v(\cdot)$: The verbalizer maps the attribute label to the corresponding tokens.
%\end{itemize}
%The evaluation result of each evaluator can be computed based on the log probability of the output sequence given the input sequence, which are determined by $g$ and $f$, respectively. 
For each aspect, we devise multiple pattern evaluators to alleviate the potential bias caused by the pattern design \cite{zhao2021calibrate}, and weight the scores of all the evaluators to obtain the final result:
%and use the ensemble of their scores to obtain the final result:
%
\begin{align}
S(I) = \sum_{j=1}^{N_E} \beta_j(I)\cdot s_j(I)
\label{eqn:weightpe}
\end{align}
where $N_E$ is the number of pattern evaluators, $S(I)$ denotes the overall score for each aspect, $\beta_j(I)$ is a factor to weight the pattern evaluators of the corresponding aspect and $s_j(I)$ indicates the score of each pattern evaluator based on the generation probability of the pre-trained model.

%In this paper, we utilize PEGASUS \cite{zhang2020pegasus} as our pre-trained model whose pre-training task is to generate the gap sentence from the remaining sentences of documents. In the next section, we will describe the evaluators for different aspects whose input and output templates are both designed based on PEGASUS's pre-training tasks.

\subsection{Evaluation Aspect}

\subsubsection{Coherence}

Coherence aims to measure whether the sentences in the generated text are semantically relevant to compose a coherent body \cite{vaku2018coherence,yuan2021bartscore}, which reflects the quality of the generated text itself.
%in the generated text is semantically relevant to the other sentences \cite{vaku2018coherence,yuan2021bartscore}.
Assume that the generated text $Y$ consists of $M$ sentences, i.e., $Y=(Y_1,Y_2,\cdots,Y_M)$, we devise $M$ pattern evaluators $E_j=(f_j,g_j)(1\leq j\leq M)$ to measure the relevance between each sentence and all the remaining sentences:
%whose input and output pattern functions are shown as follows:
\begin{align}
    f_j(I)&=Y_{\backslash j}= Y_1 \cdots Y_{j-1} \texttt{[M]} Y_{j+1}\cdots Y_M \\
    g_j(I)&=Y_j
\end{align}
where $Y_{\backslash j}$ indicates the generated text $Y$ with the $j$-th sentence replaced by a mask token $\texttt{[M]}$. 
%This design is expected to evaluate the relevance between each sentence and the remaining sentences.
%the text sequence which replaces the j-th sentence in $Y$ with a mask token $[M]$ and 
%The verbalizer $v$ is set to be empty because coherence does not measure the relationship between generated texts and attribute labels. 
The score of each pattern evaluator $E_j$ can be computed via the log probability of the pre-trained model $P_{\theta}$:
\begin{align}
    s_j(I)&=\log P_{\theta}(g_j(I)|f_j(I)) = \log P_{\theta}(Y_j|Y_{\backslash j})
\end{align}
%
%where $P_{\theta}$ denotes the pre-trained model. 

Since specific and informative sentences are more likely to impact the quality of the whole text, we adopt normalized inverse sentence frequency (NISF) \cite{zhang2018specificity} of the output sentence which can reflect its specificity to weight each pattern evaluator:
\begin{align}
    \beta_j(I)&=\textrm{NISF} (Y_j)=\frac{\textrm{ISF}(Y_j)}{\sum_{k=1}^M \textrm{ISF}(Y_k)} \\
    \textrm{ISF}(Y_j)&=\max_{w\in Y_j}\textrm{IWF}(w)
\end{align}
where the inverse sentence frequency (ISF) of $Y_j$ is computed by the maximum inverse word frequency (IWF) of the words in $Y_j$. We estimate IWF on a general corpus BookCorpus \cite{zhu2015bookcorpus}, which is commonly adopted as the pre-training dataset in the existing works \cite{devlin2019bert}:
\begin{align}
    \textrm{IWF}(w)=\frac{\log(1+|C|)}{f_w}
\end{align}
where $|C|$ indicates the total number of sentences in BookCorpus and $f_w$ denotes the number of sentences containing the word $w$.
Thus, the evaluation result of coherence can be obtained by the ensemble of the scores from all the pattern evaluators:
\begin{align}
    S_{coh}(I) =\sum_{j=1}^M \textrm{NISF}(Y_j)\cdot \log P_{\theta}(Y_j|Y_{\backslash j})
\end{align}

\subsubsection{Consistency}

Consistency aims to evaluate whether the generated text is consistent to the content prefix
%which reflects the relationship between the generated text and the content condition 
\cite{celi2020evalsurvey,rashkin2020plotmachine}. %To measure the relationship between the prefix $X$ and the generated text $Y$, 
We devise two symmetric pattern evaluators $E_{X\rightarrow Y}$ and $E_{Y\rightarrow X}$ to evaluate the consistency between the content prefix and the generated text as follows:
\begin{align}
    f_{X\rightarrow Y}(I)&=X \texttt{[M]}, g_{X\rightarrow Y}(I)=Y_{\backslash X}  \\
    f_{Y\rightarrow X}(I)&=\texttt{[M]} Y_{\backslash X}, g_{Y\rightarrow X}(I)=X
    %v_{X\rightarrow Y}&(a)=v_{Y\rightarrow X}(a)=\varnothing \notag
\end{align}
%
%where the verbalizer of these evaluators is still set to be empty because consistency does not involve the evaluation on attribute labels. 
where $Y_{\backslash X}$ denotes the remaining part of the generated text without the prefix.
Similar to coherence, we still adopt the log probability of the pre-trained model as the pattern evaluator's score and weight them with normalized inverse sentence frequency to obtain the final result of consistency:
%use normalized inverse sentence frequency to weight each evaluator. Thus, the evaluation result of consistency can be computed as follows:
\begin{align}
    %s(E_{X\rightarrow Y},X,Y,a)&=\log P_{\theta}(Y|X[M]) \notag \\
    %\beta(E_{X\rightarrow Y},X,Y,a)&=\textrm{NISF}(Y) \notag \\
    %s(E_{Y\rightarrow X},X,Y,a)&=\log P_{\theta}(X|[M]Y) \notag \\
    %\beta(E_{Y\rightarrow X},X,Y,a)&=\textrm{NISF}(X) \notag \\
    S_{cons}(I)&=\textrm{NISF}(Y_{\backslash X})  \cdot \log P_{\theta}(Y_{\backslash X}|X\texttt{[M]}) \notag \\
    & + \textrm{NISF}(X)\cdot \log P_{\theta}(X|\texttt{[M]}Y_{\backslash X})
\end{align}

\begin{table*} [!h]
\centering
\small
\setlength{\tabcolsep}{1.0mm}{
\begin{tabular}{l|c|c|c|c|c|c|c}
\toprule
%\multirow{2}*{Task}  & \multirow{2}*{\#Prefixes} & \multirow{2}*{\#Labels} & \multirow{2}*{\#Models} & \multirow{2}*{\#Samples} & \#Ratings & \multirow{2}*{Length} \\
%&   &  &   &   & (per sample) &  \\
Task  & \#Prefixes & \#Labels & \#Models & \#Samples & \#Ratings (per sample) & Length & Krippendorff's $\alpha$ \\
\midrule
Sentiment & 15 & 2 & 4 & 360 & 5 & 54.2 & 0.626 \\
Topic & 20 & 4 & 4 & 960 & 5 & 55.7 & 0.622 \\
\bottomrule
\end{tabular}}
\caption{Statistics of the evaluation set, including the number of the prefixes / attribute labels / generation models / samples / ratings (per sample), the average length of each sample and Krippendorff's $\alpha$.}
\label{tab:evalstat}
\end{table*}

\subsubsection{Attribute Relevance}

Attribute relevance aims to measure whether the generated text satisfies the attribute label \cite{da2020pplm}.
To probe the relevance between generated texts and attribute labels, we first introduce a verbalizer $v(\cdot)$ which maps all the attribute labels $a$ in the attribute set $\mathcal{A}$ to the corresponding words \cite{schick2021pet}.
Then, we design the pattern evaluators $E_j=(f_j,g_j)(1\leq j\leq N_{AR})$ where $f_j(\cdot)$ adds prompts and a mask token to the generated text, and $g_j(\cdot)$ is set to be a verbalizer:
\begin{align}
    f_j(I)&=\textrm{Concat}(\textrm{Prompt}_j,\texttt{[M]},Y) \\
    g_j(I)&=v_j(a)
\end{align}
where $\textrm{Concat}(\cdot)$ indicates the concatenation of the prompt, the mask token, and the generated text
%input text sequences 
in some order. 
%The verbalizer $v(\cdot)$ is adopted to transform the attribute label to the words. Assume that $\mathcal{A}$ denotes the set of attribute labels, 
We give an example for the pattern design of attribute relevance which is also shown in Figure \ref{fig:ctrlmodel}. In this example, the attribute is set to be the sentiment $\mathcal{A}=\{\textrm{Positive,Negative}\}$, while the patterns are designed as $f(I)=\textrm{``}Y\; \textrm{It was }\texttt{[M]}.\textrm{''}$ and $g(I)=v(\textrm{Positive/Negative})=\textrm{good/bad}$.

Inspired by the existing works \cite{schick2021pet}, we use the generation probability of the corresponding label word over all the label words as the score of the pattern evaluator:
\begin{align}
    s_j(I)=\frac{P_{\theta}(v_j(a)|f_j(I))}{\sum_{a^{'}\in \mathcal{A}} P_{\theta}(v_j(a^{'})|f_j(I))}
\end{align}
%compute the unnormalized score $s$ for each attribute label $a$ and obtain the probability distribution $q$ over all the labels with softmax:
%
%\begin{align}
    %s(X,Y,a)=\log P_{\theta}(v(a)|f(X,Y,a)) \notag \\
    %q(X,Y,a)=\frac{e^{s(X,Y,a)}}{\sum_{a^{'}\in \mathcal{A}} e^{s(X,Y,a^{'})}} \notag
%\end{align}
%

Based on the assumption that the pattern evaluator is adequate to measure the data sample if the words of all the attribute labels are easily generated, we devise the unnormalized weighted score of each evaluator as the sum of generation probabilities over all the attribute labels:
\begin{align}
    w_j(I)&=\sum_{a^{'}\in \mathcal{A}} P_{\theta}(v_j(a^{'})|f_j(I)) \\
    \beta_j(I)&=\frac{w_j(I)}{\sum_{k=1}^{N_{AR}}w_k(I)}
\end{align}

Similarly, the evaluation result of attribute relevance can be acquired by the weighted sum of all the pattern evaluators' scores:
\begin{align}
    S_{AR}(I)=\sum_{j=1}^{N_{AR}} \beta_j(I) \cdot s_j(I)
\end{align}

% TODO: 强调分数结果是[0,1]的数，物理意义是生成文本的符合程度，而不是标签类别

% BARTscore的结果可能需要update，原因是replace

\section{Experiment}

\subsection{Datasets}
\label{sec:dataset}

Since there is no standard benchmark dataset for evaluating controlled text generation, we construct an evaluation set to measure the correlation between automatic metrics and human judgments.

\noindent \textbf{Task}: We choose sentiment-controlled and topic-controlled text generation as the benchmark tasks, which are widely used in the existing works \cite{da2020pplm,chan2021cocon}. These two tasks require the models to generate texts conditioned on the given prefixes and sentiment / topic labels, respectively. In the task of sentiment-controlled text generation, we follow PPLM \cite{da2020pplm} and CoCon \cite{chan2021cocon} to adopt 15 prefixes and 2 sentiment labels (i.e., positive and negative). As for topic-controlled text generation, we follow CoCon \cite{chan2021cocon} to adopt 20 prefixes and 4 topic labels (i.e., computers, politics, religion, and science).

\noindent \textbf{Generation Models}:
We consider various generation models including CTRL \cite{keskar2019ctrl}, PPLM \cite{da2020pplm}, GeDi \cite{krause2020gedi}, and CoCon \cite{chan2021cocon}. These representative models support both the sentiment-controlled and topic-controlled text generation tasks,
%the tasks of text generation with controlled sentiments / topics
and cover different levels of generation abilities. We make these models generate 3 different samples for each unique pair of prefixes and attribute labels. We set the maximum length of generated texts to be 80 and remove the last sentence if it is not complete. We directly use the generation results if they have been released by the original papers. Otherwise, we run the original codes to obtain the generation results.

\noindent \textbf{Human Annotation}: We collect human ratings on the generated texts from Amazon Mechanical Turk (AMT). Each survey of AMT contains a prefix, an attribute label, and five generated texts including (a) four generated texts from the above four models respectively, and (b) one negative sample which is constructed by perturbing (e.g. sentence shuffling and dropping) another sample from the evaluation set \cite{guan2021openmeva}. We ask annotators to 
%compare the quality of five texts 
%based on some aspect (such as coherence, consistency or attribute relevance), 
%and 
rate these texts with a 1-5 Likert scale for each aspect. To control the annotation quality, we discard the submissions if the annotator assigns a higher rating to the negative sample than other texts. We ensure that each generated text contains 5 valid ratings for each aspect, where the average value of valid ratings is used as the human judgments. We also calculate Krippendorff's $\alpha$ \cite{krippendorff2018alpha} to show the agreement of human ratings, which is 0.626 / 0.622 for sentiment-controlled / topic-controlled text generation tasks, respectively.
%As for the agreement of human ratings, the percentage of data with at least 3/5 (4/5) same ratings is 93.5\% (53.8\%) in the sentiment-controlled text generation task. While in the topic-controlled text generation task, the percentage is 93.1\% (52.7\%) for the data with at least 3/5 (4/5) same ratings.

The statistics of the evaluation set are shown in Table \ref{tab:evalstat}.

\subsection{Implementation Details}

\begin{table} [!t]
\centering
\scriptsize
\setlength{\tabcolsep}{1.0mm}{
\begin{tabular}{l|c|c|c|c}
\toprule
Task  & \#Seed Prompts & \#Prompts & \#Verbalizers & \#Evaluators \\
\midrule
Sentiment & 3 & 24 & 3 & 72 \\
Topic & 4 & 32 & 1 & 32 \\
\bottomrule
\end{tabular}}
\caption{Statistics of the pattern evaluators in attribute relevance. The number of evaluators is obtained by multiplying the number of prompts and verbalizers.}
\label{tab:promptstat}
\end{table}

We choose PEGASUS \cite{zhang2020pegasus} as our base model in the overall result and also explore other pre-trained models in \S\ref{sec:basemodel}. The hyper-parameters of Transformer blocks are the same as PEGASUS-large with 568M parameters. As for the pattern evaluators in attribute relevance involving prompts and verbalizers which need to be additionally designed,
%As for the prompt design in the evaluation of controllability, 
we follow BARTScore \cite{yuan2021bartscore} to first adopt manually devised seed prompts and verbalizers in the existing works \cite{schick2021pet,schick2020petforgen}, and then collect paraphrases to automatically expand our evaluator set. The statistics of pattern evaluators in attribute relevance are presented in Table \ref{tab:promptstat}. More details about the specific design of 
prompts and verbalizers are included in Appendix \ref{app:evaluator}.

\begin{table*} [!h]
\centering
\small
\setlength{\tabcolsep}{0.9mm}{
\begin{tabular}{l|c|c|c|c|c|c|c|c|c|c|c|c}
\toprule
Task & \multicolumn{6}{c|}{Sentiment} & \multicolumn{6}{c}{Topic} \\
\midrule
Aspect  & \multicolumn{3}{c|}{Coherence} & \multicolumn{3}{c|}{Consistency} & \multicolumn{3}{c|}{Coherence} & \multicolumn{3}{c}{Consistency}  \\
\cmidrule{1-13}
Metric & $r$ & $\rho$ & $\tau$ & $r$ & $\rho$ & $\tau$ & $r$ & $\rho$ & $\tau$ & $r$ & $\rho$ & $\tau$ \\
\midrule
DisScore & 0.2938 & 0.2329 & 0.1664  & 0.2010 & 0.1662 & 0.1178 & 0.1526  &  0.1315 & 0.0937  & 0.0053 & 0.0072 & 0.0051  \\
UNION & 0.2317 & 0.2571 & 0.1836 & 0.1925 & 0.1422 & 0.1009 & 0.1628 & 0.1300 & 0.0924 & 0.0664 & 0.0777 & 0.0553 \\
BLEURT & 0.2585 & 0.2606 & 0.1850 & 0.2382  & 0.2012  & 0.1445 & 0.1631 & 0.1428 & 0.1016 & 0.0433 & 0.0607 & 0.0443 \\
\midrule
PPL-GPT & 0.3376 & 0.3310 & 0.2350  & 0.1881 & 0.1672 & 0.1203 & 0.1459 & 0.1316 & 0.0940 & 0.1013 & 0.0841 & 0.0595 \\
PPL-PEGASUS & 0.3901 & 0.3860 & 0.2743 & 0.2728 & 0.2513 & 0.1808  & 0.1420 & 0.1313 & 0.0929 & 0.1883 & 0.1771 & 0.1235 \\
BARTScore & 0.3880  & 0.3848 & 0.2736  & 0.2682 & 0.2533 & 0.1804 & 0.1599 & 0.1325 & 0.0939 & 0.1528 & 0.1408 & 0.0978 \\
BARTScore-PEGASUS & 0.3853  & 0.3712  &  0.2653  & 0.2480  & 0.2267  & 0.1630  & 0.1638  & 0.1493  & 0.1048  & 0.1539 & 0.1362  & 0.0953  \\
\midrule
%\hline
CTRLEval (Ours) & \textbf{0.4395} & \textbf{0.4208} & \textbf{0.3044}  & \textbf{0.3226} & \textbf{0.3096} & \textbf{0.2235} & \textbf{0.2403} & \textbf{0.2245} & \textbf{0.1582} & \textbf{0.2342} & \textbf{0.2281} & \textbf{0.1595} \\
\bottomrule
\end{tabular}}
\caption{Pearson ($r$), Spearman ($\rho$), and Kendall ($\tau$) correlations of coherence and consistency in sentiment-controlled and topic-controlled text generation. }
\label{tab:maincohcons}
\end{table*}

\begin{table} [!h]
\centering
\scriptsize
\setlength{\tabcolsep}{0.55mm}{
\begin{tabular}{l|c|c|c|c|c|c}
\toprule
Task & \multicolumn{3}{c|}{Sentiment} & \multicolumn{3}{c}{Topic} \\
\midrule
Aspect  & \multicolumn{3}{c|}{Attr. Rel.} & \multicolumn{3}{c}{Attr. Rel.}  \\
\cmidrule{1-7}
Metric & $r$ & $\rho$ & $\tau$ & $r$ & $\rho$ & $\tau$ \\
\midrule
DisScore & 0.2213 & 0.2914 & 0.2068 & 0.3624 & 0.2777 & 0.1969 \\
UNION & -0.0133 & -0.0324 & -0.0219 & -0.0483 & -0.0635 & -0.0455  \\
BLEURT & 0.0801 & 0.0652 & 0.0467 & 0.1040 & 0.0841 & 0.0604 \\
\midrule
PPL-GPT & -0.0197 & -0.0472 & -0.0338 & 0.0853 & 0.1084 & 0.0769 \\
PPL-PEGASUS & 0.0356 & -0.0070 & -0.0083 & 0.0611 & 0.0662 & 0.0480 \\
BARTScore & -0.0006 & -0.0488 & -0.0372  & 0.0776 & 0.0853 & 0.0603 \\
BARTScore-PEGASUS & 0.0336 & -0.0271  & -0.0221  &  0.0605 & 0.0567  &  0.0402 \\
\midrule
%\hline
CTRLEval (Ours) & \textbf{0.2861} & \textbf{0.3008} & \textbf{0.2111}  & \textbf{0.5189} & \textbf{0.4006} & \textbf{0.2865} \\
\bottomrule
\end{tabular}}
\caption{Pearson ($r$), Spearman ($\rho$), and Kendall ($\tau$) correlations of attribute relevance in sentiment-controlled and topic-controlled text generation. Note that the baselines which are not trained on attribute-annotated corpora can hardly measure the relevance between generated texts and attribute labels, thereby causing low correlations.}
\label{tab:maincont}
\end{table}

\subsection{Baselines}

We choose several state-of-the-art reference-free metrics as our baselines:

\noindent \textbf{Perplexity (PPL)} \cite{brown1992perplexity}: This method calculates the perplexity of generated texts with a language model. We use GPT \cite{radford2018gpt}
%as the base model. We additionally use PEGASUS \cite{zhang2020pegasus} for a fair comparison.
and PEGASUS \cite{zhang2020pegasus} as the base models since GPT is commonly used in the existing works \cite{da2020pplm} and PEGASUS is our base model. They are denoted as \textbf{PPL-GPT} and \textbf{PPL-PEGASUS}, respectively.

%\noindent \textbf{RUBER-BERT} \cite{sarik2019ruberbert}: This method is a self-supervised baseline which is trained to distinguish the hu

\noindent \textbf{Discriminator Score (DisScore)} \cite{kannan2017disscore,chan2021cocon}: This method trains a discriminator with different objectives. We adopt the IMDB movie review dataset \cite{mass2011imdb} / HuffPost News category dataset\footnote{\url{https://www.kaggle.com/rmisra/news-category-dataset}} \cite{misra2018news} for sentiment-controlled / topic-controlled text generation tasks, respectively. For coherence and consistency, the discriminator is trained to distinguish human-written texts from manually constructed negative samples, where the ratio of positive and negative samples is 1:1. For attribute relevance, it is trained based on the sentiment / topic classification task, respectively \cite{chan2021cocon}. Both the sentiment and topic discriminators are implemented based on BERT \cite{devlin2019bert} and they achieve 94.15\% / 91.54\% on the corresponding test set, respectively.

%For coherence and consistency, the discriminator is trained to distinguish human-written texts from manually constructed negative samples. For attribute relevance, it is trained on the IMDB movie review dataset (2-class) \cite{mass2011imdb} / HuffPost News category dataset\footnote{\url{https://www.kaggle.com/rmisra/news-category-dataset}} (4-class) for sentiment-controlled / topic-controlled text generation tasks, respectively \cite{chan2021cocon}. Both the sentiment and topic discriminators are implemented based on BERT \cite{devlin2019bert} and they achieve 94.15\% / 91.54\% on the corresponding test set, respectively.

%We followed the existing work \cite{chan2021cocon} to train a sentiment / topic discriminator on the IMDB movie review dataset (2-class) \cite{mass2011imdb} / HUff News category dataset\footnote{\url{https://www.kaggle.com/rmisra/news-category-dataset}} (4-class), and used the prediction probability on the given attribute label. Both of the sentiment and topic discriminators are implemented based on BERT \cite{devlin2019bert} and they achieve xx.xx\% / 91.54\% on the corresponding test set, respectively. 

\noindent \textbf{UNION} \cite{guan2020union}: This method is a self-supervised metric which is trained to distinguish human-written texts from the automatically perturbed negative samples with well-designed negative sampling strategies and multi-task learning.
We use the same datasets as the discriminator score to train UNION.

% TODO: 确定一下BLEURT原文是否需要reference
\noindent \textbf{BLEURT} \cite{sellam2020bleurt}: This method is a supervised metric which is pre-trained on synthetic examples and then fine-tuned to fit human ratings. We used the same instruction in \S\ref{sec:dataset} to additionally annotate the generated texts to construct the training set for BLEURT, whose amount is the same as the evaluation set. There is no overlap between BLEURT's training set and the evaluation set.

\noindent \textbf{BARTScore} \cite{yuan2021bartscore}: This method utilizes the generation probabilities of BART \cite{lewis2020bart} to measure the relationship among sources, hypotheses, and references. Since this metric simultaneously contains referenced and reference-free parts, we only use the reference-free score in our experiments. We also use PEGASUS \cite{zhang2020pegasus} as the base model for a fair comparison, which is denoted as \textbf{BARTScore-PEGASUS}.

\subsection{Overall Result}

We follow the existing work \cite{guan2020union,yuan2021bartscore} to adopt Pearson ($r$), Spearman ($\rho$), and Kendall ($\tau$) correlation coefficients between automatic metrics and human judgments to measure the performance of different metrics. 

The overall results on sentiment-controlled and topic-controlled text generation are shown in Table \ref{tab:maincohcons} and \ref{tab:maincont}. We can observe that CTRLEval outperforms other baselines with a large margin, indicating the effectiveness of our metric on different evaluation aspects. In Table \ref{tab:maincont}, unsupervised baselines can hardly measure the relevance between generated texts and attribute labels because 
%they only provide a score without the interpretability of specific evaluation aspects.
they only provide a task-agnostic score which is weakly relevant to this specific
aspect.
For comparison, our metric, which supports the evaluation for different aspects of generated texts via the design of text infilling tasks, can obtain much better performance and even outperform the supervised baselines.
%From Table \ref{tab:maincohcons}, we can also see that the learnable metrics (such as UNION and BLEURT) which are trained on the task-specific downstream datasets may cause over-fitting and degrade their performance. For comparison, CTRLEval as an unsupervised metric can successfully generalize to different criteria and achieve promising results.

\subsection{Ablation Study}

\begin{table} [!h]
\centering
\small
\setlength{\tabcolsep}{1.0mm}{
\begin{tabular}{l|c|c|c}
\toprule
\multirow{2}*{Metric}  & \multicolumn{3}{c}{Aspect}  \\
\cmidrule{2-4}
 & Coherence & Consistency & Attr. Rel. \\
\midrule
CTRLEval (Ours) & \textbf{0.2403} & \textbf{0.2342} & \textbf{0.5189} \\
\midrule
 \multicolumn{4}{c}{Weight of Pattern Evaluators (w/o $\beta$)} \\
\midrule
w/ $\textrm{mean}(\cdot)$ & 0.2295 & 0.1927 & 0.5091 \\
w/ $\textrm{max}(\cdot)$ & 0.2323 & 0.1772 & 0.5170  \\
w/ $\textrm{min}(\cdot)$ & 0.1518 & 0.1559 & 0.4153  \\
\midrule
 \multicolumn{4}{c}{Pattern Function (w/o $f\& g$)} \\
\midrule
w/ PPL-GPT-PF & 0.2041  & 0.2169  & 0.4376  \\
w/ BARTScore-PF & 0.1236 & 0.1843 & 0.3972 \\
\bottomrule
\end{tabular}}
\caption{Pearson correlation of ablation models in topic-controlled text generation.}
\label{tab:ablation}
\end{table}

To further investigate the effect of each module, we conduct ablation studies on the weight of pattern evaluators and the design of pattern functions. For the weight of evaluators, we use the mean, maximum and minimum values of all the evaluators as the final result rather than the weighted sum based on the factor $\beta$.
%replace the coefficient $\beta$ with the common functions including the mean, maximum and minimum values to show the effectiveness of our ensemble method. 
As for the design of pattern functions, we fix the base model and replace our input and output patterns ($f\& g$) with those of PPL-GPT \cite{radford2018gpt} and BARTScore \cite{yuan2021bartscore}. The pattern functions of these ablation models are not designed for text infilling tasks. Both of them remove the mask token in the input pattern, and PPL-GPT additionally places the input pattern at the beginning of the output pattern.

%The patterns of BARTScore are similar to our method except that it does not use mask tokens in the input pattern. 

%directly uses the conditional generation probability without prompts and mask tokens.

%PPL- adopts all the information as a input sequence of the decoder and BARTScore directly uses the conditional generation probability without prompts and mask tokens.

The results in Table \ref{tab:ablation} show that each module in our metric contributes to the final performance. As for the weight of evaluators, we can observe that our weight factor performs better than common aggregation functions especially in consistency, indicating the necessity of the well-designed ensemble method when the number of pattern evaluators is small. Also, our pattern functions outperform those of other baselines, thereby showing the effectiveness of text infilling tasks which can fully utilize pre-trained models in an unsupervised setting.

%our framework which formulates the evaluation aspect into text infilling tasks.

\subsection{Analysis on Generalization Ability}

Generalization ability is essential for automatic metrics to evaluate open-ended text generation models. In this section, we will test whether our metric can be generalizable to measure the generated texts faced with model drift and quality drift.
%of different models and with different qualities.

\subsubsection{Model Drift}

\begin{figure}[!htp]
  \centering
  \includegraphics[width=1.0\linewidth]{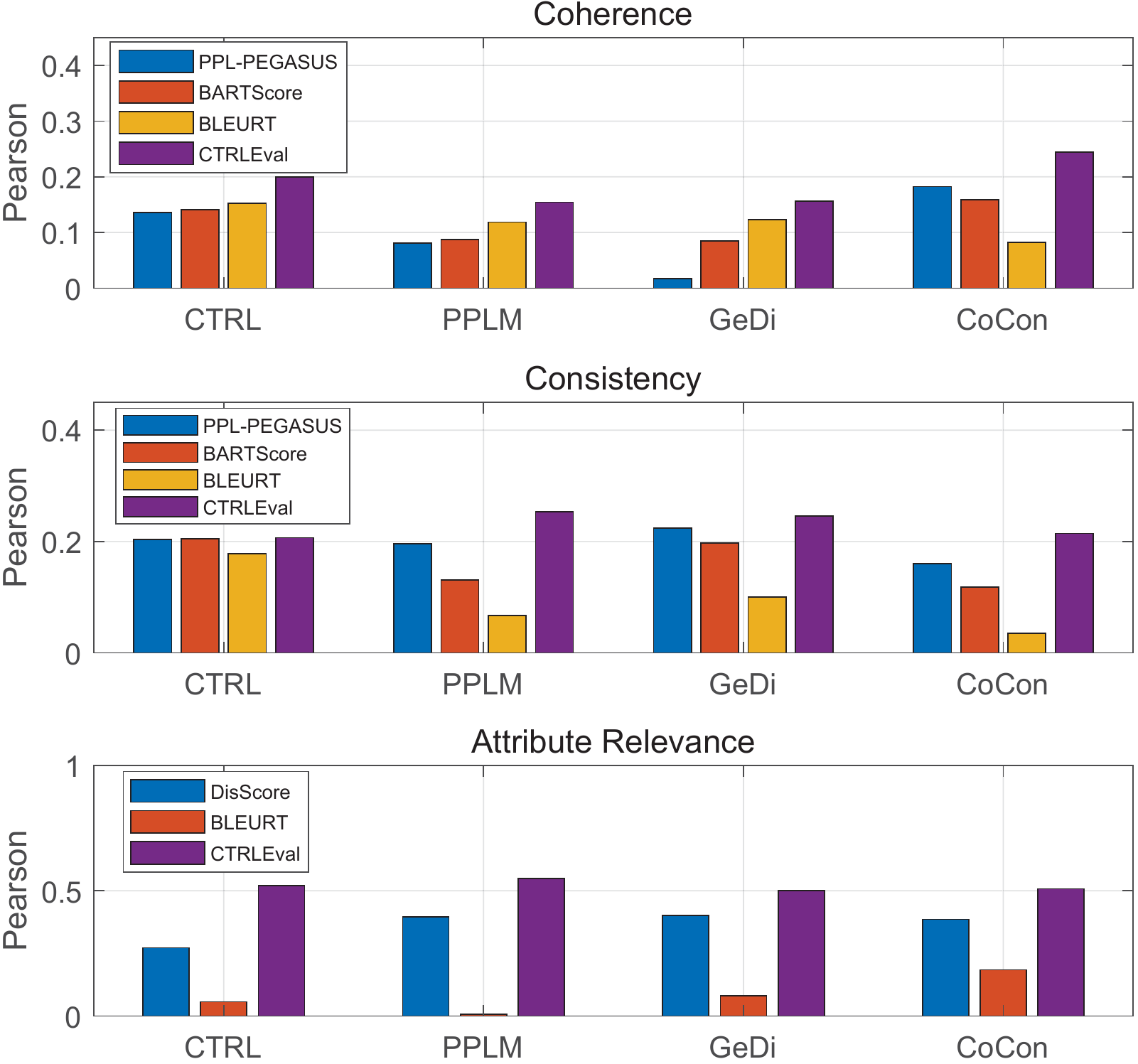} % 1\linewidth
  \caption{Pearson correlation on the generated results from four generation models in the task of topic-controlled text generation.}
  \label{fig:modeldrift}
\end{figure}

To measure whether CTRLEval is reliable to assess the generated results of different models, we split the evaluation set into four subsets based on the generation model and calculate Pearson correlation between each metric and human judgments. 

The results in Figure \ref{fig:modeldrift} show that our metric can outperform other baselines on the generated texts of all the generation models. Simultaneously, CTRLEval can achieve stable performance with smaller variances when evaluating different generation models, indicating that our metric can generalize to the model drift better. 
%For comparison, there is commonly a significant performance gap for the baseline metrics to evaluate different models, such as PPL-GPT's evaluation results on CTRL and PPLM.

\subsubsection{Quality Drift}

\begin{figure}[!htp]
  \centering
  \includegraphics[width=1.0\linewidth]{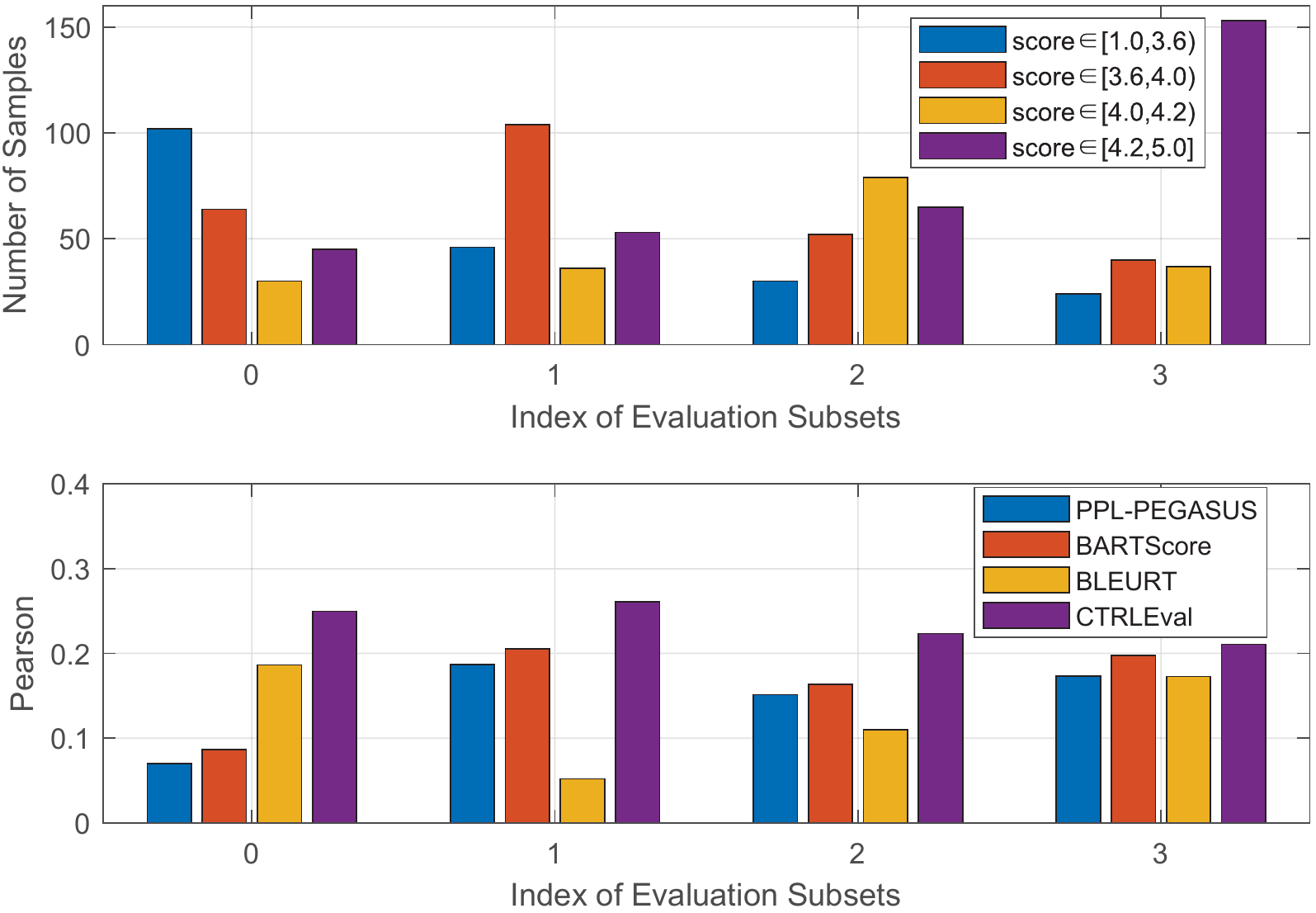}% 1\linewidth
  \caption{\textbf{Top}: The number of samples with different coherence scores in the four biased evaluation subsets. \textbf{Bottom}: Pearson correlation of different metrics on the biased evaluation subsets.}
  \label{fig:qualitydrift}
\end{figure}

To evaluate the generalization ability of CTRLEval on the generated texts with different qualities, we follow the existing work \cite{sellam2020bleurt,guan2020union} to construct four biased subsets based on the coherence score of topic-controlled text generation. We first sort all the samples in the evaluation set and use the quartiles to split them into four subsets with the index from 0 to 3.
%split all the samples in the evaluation set into four subsets based on their human scores.
Then, we create four biased subsets. For the $j^{th}$ subset, we sampled the generated texts which belong to the original $i^{th}$ subset with a probability of $\frac{1}{|j-i|+1}$ where $i,j=0,1,2,3$. Thus, the four biased subsets have different distributions of generated texts with different qualities, as shown in Figure \ref{fig:qualitydrift}.

%The statistics of each set are shown in Figure \ref{fig:qualitydrift}. Intuitively, the quality of the generated texts in each subset is better when the index is larger.

We then calculate the Pearson correlation between each metric and human judgments. The results in Figure \ref{fig:qualitydrift} show that CTRLEval has higher correlations than the baselines on the evaluation subsets with different qualities. 
%Also, our metric can be more stable on the subsets of different qualities, 
Also, our metric can achieve more stable performance on different subsets,
which shows our better generalization ability to deal with quality drift.

\subsection{Analysis on the Number of Evaluators}

\begin{figure}[!htp]
  \centering
  \includegraphics[width=1.0\linewidth]{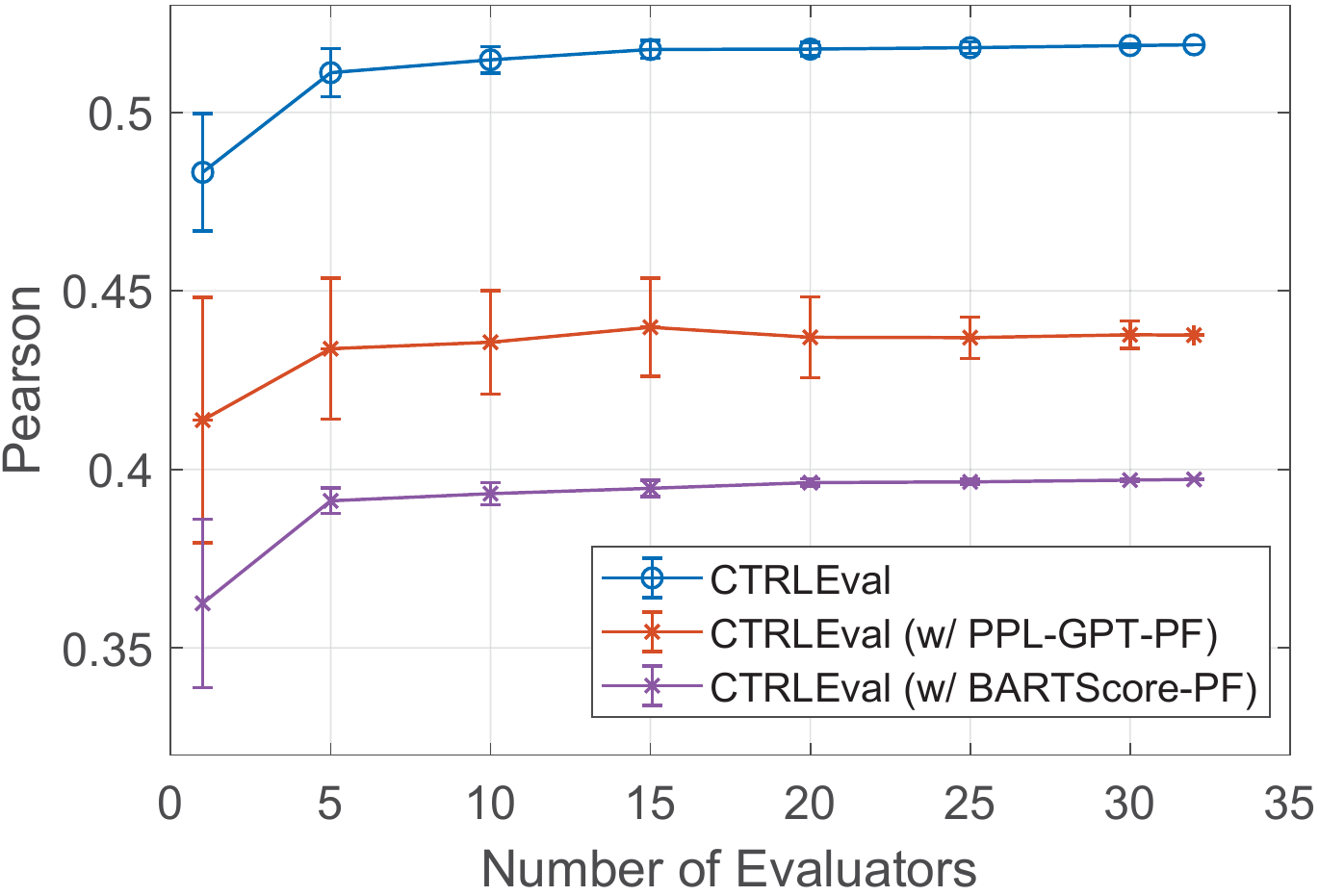} % 1\linewidth
  \caption{Pearson correlation of the models with different numbers of evaluators.}
  %attribute relevance in the task of topic-controlled text generation.}
  \label{fig:numbereval}
\end{figure}

To investigate how the number of pattern evaluators affects the performance, we randomly sample the evaluators 20 times when evaluating attribute relevance in topic-controlled text generation,
and illustrate mean values and standard deviations of each number of evaluators in Figure \ref{fig:numbereval}.

Figure \ref{fig:numbereval} shows that as the number of evaluators increases, the mean value of our performance can be persistently improved while the standard deviation is gradually reduced. This demonstrates the necessity of devising multiple pattern evaluators for each aspect, which can alleviate the bias brought by the pattern design. The comparison between the pattern functions of CTRLEval and other baselines indicates our superior performance on all the numbers of evaluators.

\subsection{Analysis on Base Model}
\label{sec:basemodel}

\begin{table} [!h]
\centering
\small
\setlength{\tabcolsep}{1.0mm}{
\begin{tabular}{l|c|c|c|c}
\toprule
\multirow{2}*{Base Model} & \multirow{2}*{\#Param} & \multicolumn{3}{c}{Aspect}  \\
\cmidrule{3-5}
 & & Coherence & Consistency & Attr. Rel. \\
\midrule
PEGASUS & 568M & 0.3044 & 0.2235 & \textbf{0.2111} \\
BART & 400M & \textbf{0.3123} & 0.1650 &  0.1951 \\
T5 & 770M & 0.2930 & \textbf{0.2350} & 0.2075  \\
\bottomrule
\end{tabular}}
\caption{Kendall correlation of CTRLEval with different base models in sentiment-controlled text generation. \#Param means the number of parameters.}
\label{tab:basemodel}
\end{table}

Since our method can adapt to different pre-trained models whose pre-training task is text infilling, we additionally choose BART \cite{lewis2020bart} and T5 \cite{raffel2020t5} as our base model, and present the results in Table \ref{tab:basemodel}.

Table \ref{tab:basemodel} shows that PEGASUS and T5 obtain comparable performance on all the evaluation aspects, which indicates that our well-designed text infilling tasks can be transferable to T5 without considerable modification. As for BART which performs worse on consistency and attribute relevance, we conjecture that the fewer parameters and the form of pre-training tasks may limit the performance. Since the pre-training task of BART is to generate the complete text rather than only the masked part of the input text, 
it may not be good at the evaluation involving a short span of texts, such as the prefix in the evaluation of consistency and the label word in attribute relevance.
%it may lack the ability to measure the relationship among different parts of generated texts, such as the prefix and the remaining text in the evaluation of consistency.

%there are two possible reasons: 1) BART has fewer parameters which may limit the performance. 2) The pre-training task of BART requires the model to generate the complete text rather than only the masked part, which is slightly different from our design criterion of text infilling tasks. This gap may affect the performance since we use the base model in an unsupervised setting.

%As for consistency and attribute relevance, BART performs worse than other two models. We conjecture that there are two possible reasons: 1) BART has fewer parameters which may degrade the performance. 2) The pre-training task of BART is slightly different from PEGASUS and T5, which requires the model to generate the complete text rather than only the masked part

We also provide the analysis on the number of parameters in Appendix \ref{app:numberofparam} and the case study in Appendix \ref{app:casestudy}.

\section{Discussion}

\noindent \textbf{Extension to More Control Variables}: In this paper, we evaluate the relationship between generated texts and two control variables (including content prefixes and attribute labels) via consistency and attribute relevance, respectively. We can also extend our metric to other control variables by designing additional pattern evaluators to measure the relationship between generated texts and each variable, respectively. We will further investigate the extensibility of our metric in the future work.

\noindent \textbf{Design of Pattern Evaluators}: With the rapid development of prompt tuning, recent works have proposed new methods on the design of prompts and verbalizers \cite{gao2021make,lester2021prompttuning}, which provide alternatives to our metric in attribute relevance. Also, the weight factor of each evaluator can be set as diversity metrics \cite{hashimoto2019unify} besides NISF in coherence and consistency. We will leave the exploration of more settings on pattern evaluators as the future work.

%We devise specific prompts and verbalizers in attribute relevance according to the existing works on pattern-exploiting training \cite{schick2020petforgen,schick2021pet} and prompt tuning \cite{gu2021ppt,lester2021prompttuning}. And we will also explore new methods on the design of prompts and verbalizers \cite{gao2021make} which are proposed recently in the future work. 

%Since this research direction is active recently, many research works discuss how to improve the design of prompts and verbalizers \cite{gao2021make}. believe that our metric can directly benefit from the development of this direction and  

%Recent works on the design of prompts and verbalizers have emerged in this active research direction \cite{gao2021make}. Thus, we believe that our metric can benefit from the development

\section{Conclusion}

We present an unsupervised reference-free metric called CTRLEval for evaluating controlled text generation. This metric formulates the evaluation of different aspects into multiple text infilling tasks, and utilizes the ensemble of generation probabilities from a pre-trained model in different tasks as the evaluation result. Experimental results indicate that CTRLEval obtains higher correlations with human judgments and shows better generalization ability for addressing model drift and quality drift.
%%evaluating generated texts from different models or with different qualities.

\section*{Acknowledgments}

This work was supported by the National Science Foundation for Distinguished Young Scholars (with No. 62125604) and the NSFC projects (Key project with No. 61936010 and regular project with No. 61876096). This work was also supported by the Guoqiang Institute of Tsinghua University, with Grant No. 2019GQG1 and 2020GQG0005.

\section*{Ethics Statement}

We construct an evaluation set for evaluating controlled text generation. The data samples in this set are all from the existing works with open-source codes, model checkpoints, and generated results. We directly use the generated results if the authors have released them. Otherwise, we adopt the same setting as the original papers to make these models generate texts. We do not apply extra selection strategies to the generated results. 

We resort to Amazon Mechanical Turk (AMT) for the annotation of this evaluation set. We do not invade the privacy or collect personal information of annotators. We pay each annotator \$0.06 for each survey including four generated texts and one negative sample. The payment is determined based on the length of data samples. We additionally ask annotators to check whether there is a potential ethical problem in the data, and remove these problematic data in the evaluation set. After annotation on AMT, we manually review all the annotated samples from an ethical perspective. However, we admit that there may still exist unpredictable bias in this evaluation set.

\bibliography{custom}
\bibliographystyle{acl_natbib}

%\newpage
\appendix

\section{Pattern Evaluator for Attribute Relevance}
\label{app:evaluator}

We first choose the prompts and verbalizers which have been shown to work well in the existing works on few-shot text classification \cite{schick2021pet,gao2021make} and generation \cite{schick2020petforgen} as the seed prompts and verbalizers. Then, we expand our prompt set with the following rules: 1) Switching the order of generated texts, prompts, and mask tokens; 2) Collecting the paraphrases of seed prompts just as BARTScore \cite{yuan2021bartscore} does. All the prompts and verbalizers which are used in our experiments are shown in Table \ref{tab:promptverb}.

\section{Analysis on the Number of Parameters}
\label{app:numberofparam}

\begin{table} [!h]
\centering
\small
\setlength{\tabcolsep}{1.0mm}{
\begin{tabular}{l|c|c|c|c}
\toprule
\multirow{2}*{Base Model} & \multirow{2}*{\#Param} & \multicolumn{3}{c}{Aspect}  \\
\cmidrule{3-5}
 & & Coherence & Consistency & Attr. Rel. \\
\midrule
T5-small & 60M & 0.2389 & 0.1495 &  0.1765 \\
T5-base & 220M & 0.2847 &  0.2053 & 0.1867  \\
T5-large & 770M & \textbf{0.2930} & \textbf{0.2350} & \textbf{0.2075}  \\
\bottomrule
\end{tabular}}
\caption{Kendall correlation of CTRLEval with T5-small, T5-base, and T5-large in sentiment-controlled text generation. \#Param means the number of parameters.}
\label{tab:basemodelparam}
\end{table}

We further conduct experiments on the base model with different numbers of parameters. Since the authors of PEGASUS \cite{zhang2020pegasus} do not release the model checkpoint of PEGASUS-base, we choose T5-small, T5-base and T5-large \cite{raffel2020t5} as our base models respectively, and present the results in Table \ref{tab:basemodelparam}. The results show that larger numbers of parameters can benefit the model performance while degrading the computation efficiency.

\section{Case Study}
\label{app:casestudy}

\begin{figure}[!t]
  \centering
  \includegraphics[width=1.0\linewidth]{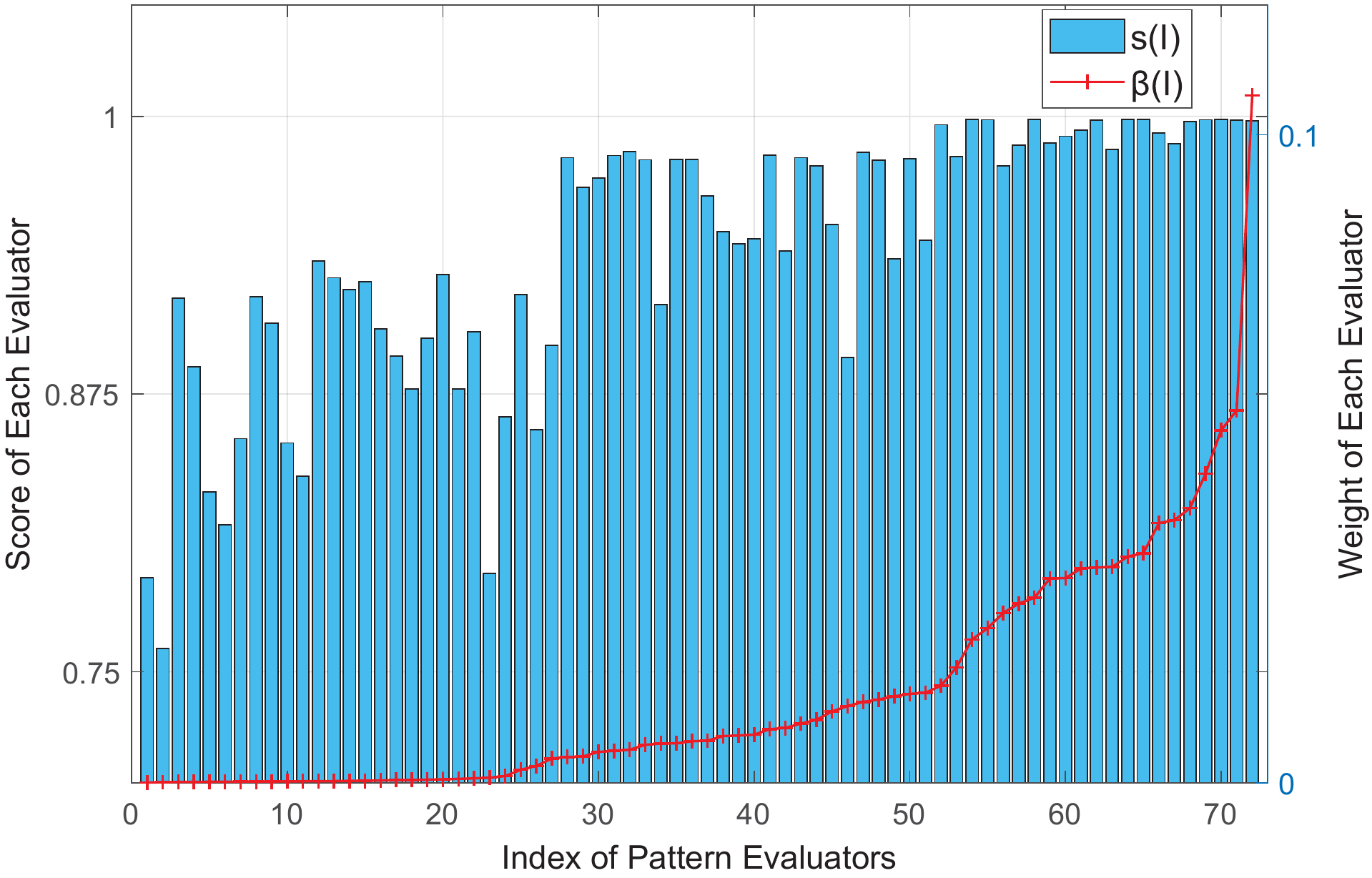} % 1\linewidth
  \caption{The score and weight of each evaluator for evaluating attribute relevance of the second sample in Figure \ref{fig:casestudycont}.}
  \label{fig:casestudyspecific}
\end{figure}

\begin{figure*}[!t]
  \centering
  \includegraphics[width=1.0\linewidth]{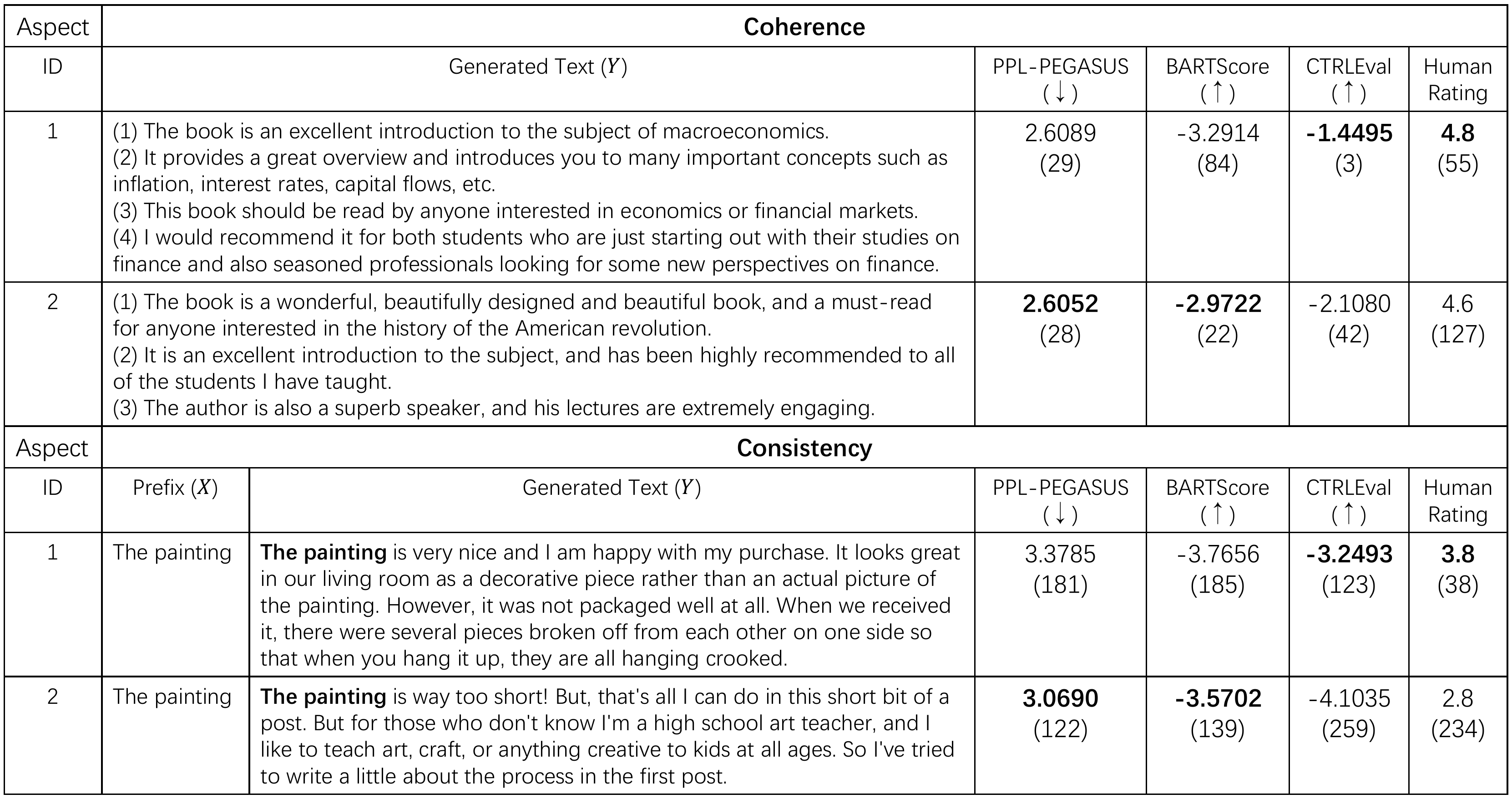} % 1\linewidth
  \caption{Evaluation cases on coherence and consistency in sentiment-controlled text generation. The result of each metric is provided by the absolute value of the evaluation score and the rank of this score over all the evaluation results of the corresponding metric. To improve readability, we label each sentence with its order in the evaluation of coherence and highlight the prefix in bold when evaluating consistency.}
  \label{fig:casestudycohcons}
\end{figure*}

\begin{figure*}[!t]
  \centering
  \includegraphics[width=1.0\linewidth]{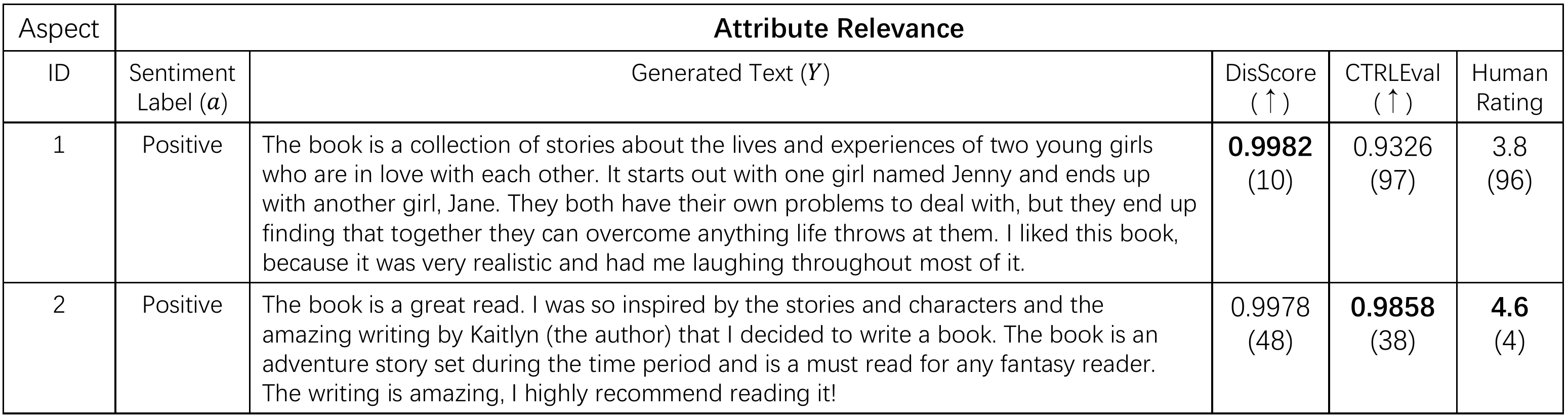} % 1\linewidth
  \caption{Evaluation cases on attribute relevance in sentiment-controlled text generation. The result of each metric is provided by the absolute value of the evaluation score and the rank of this score over all the evaluation results of the corresponding metric.}
  \label{fig:casestudycont}
\end{figure*}

To intuitively show how our metric works in the evaluation of controlled text generation, we provide some cases on the three evaluation aspects, including coherence (Figure \ref{fig:casestudycohcons}), consistency (Figure \ref{fig:casestudycohcons}), and attribute relevance (Figure \ref{fig:casestudycont}). Since the range of various metrics is always different, it may be less meaningful to directly compare the absolute value of each metric. Thus, we follow the existing works \cite{guan2020union,liu2021mars} to conduct a pairwise comparison on different samples.

The results in Figure \ref{fig:casestudycohcons} and \ref{fig:casestudycont} show that our metric can give accordant preferences with human judgments, indicating the effectiveness of our metric on all three evaluation aspects. To further show how each pattern evaluator works in the overall evaluation result, we take the second sample in Figure \ref{fig:casestudycont} as an example and visualize the weight $\beta(I)$ and score $s(I)$ in Figure \ref{fig:casestudyspecific}. We can observe that most of the pattern evaluators assign high scores to this sample which agree with the human judgment. Simultaneously, the weight factor automatically reduces the effect of low-quality evaluators which also plays an important role in the final evaluation result.

\begin{table*} [!t]
\centering
\small
\setlength{\tabcolsep}{1.0mm}{
\begin{tabular}{l|l|l}
\toprule
Task & \multicolumn{2}{l}{Sentiment} \\
\midrule
\multirow{8}*{$f(I)$} & Seed Prompt & Expanded Prompt \\
\cmidrule{2-3}
  & \multirow{3}*{$Y$ In summary, it was $\texttt{[M]}$.} & In summary, it was $\texttt{[M]}$. $Y$ \quad $Y$ To sum up, it was $\texttt{[M]}$. \quad To sum up, it was $\texttt{[M]}$. $Y$ \\
  & & $Y$ All in all, it was $\texttt{[M]}$. \qquad All in all, it was $\texttt{[M]}$. $Y$ \qquad $Y$ In brief, it was $\texttt{[M]}$. \\
  & & In brief, it was $\texttt{[M]}$. $Y$ \\
\cmidrule{2-3}
  & \multirow{2}*{$Y$ It was $\texttt{[M]}$.} & It was $\texttt{[M]}$. $Y$ \qquad $Y$ It seems $\texttt{[M]}$. \qquad It seems $\texttt{[M]}$. $Y$ \qquad $Y$ It appears $\texttt{[M]}$. \\
  & & It appears $\texttt{[M]}$. $Y$ \quad $Y$ It becomes $\texttt{[M]}$. \quad It becomes $\texttt{[M]}$. $Y$ \\
  \cmidrule{2-3}
  & \multirow{2}*{$Y$ Really $\texttt{[M]}$!} & Really $\texttt{[M]}$! $Y$ \qquad $Y$ Just $\texttt{[M]}$! \qquad Just $\texttt{[M]}$! $Y$ \qquad $Y$ Actually $\texttt{[M]}$!  \\
  &    &  Actually $\texttt{[M]}$! $Y$ \quad $Y$ So $\texttt{[M]}$! \qquad So $\texttt{[M]}$! $Y$ \\
\midrule
\multirow{2}*{$g(I)$} & \multicolumn{2}{l}{Verbalizer} \\
\cmidrule{2-3}
 & \multicolumn{2}{l}{$v(\textrm{Positive}, \textrm{Negative})=\{(\textrm{good}, \textrm{bad}), (\textrm{positive},\textrm{negative}), (\textrm{great}, \textrm{terrible})\}$} \\
 \midrule
 \midrule
 Task & \multicolumn{2}{l}{Topic} \\
 \midrule
\multirow{11}*{$f(I)$} & Seed Prompt & Expanded Prompt \\
\cmidrule{2-3}
 &  \multirow{2}*{$Y$ News: $\texttt{[M]}$} &  News: $\texttt{[M]}$ $Y$ \qquad $Y$ Article: $\texttt{[M]}$ \quad Article: $\texttt{[M]}$ $Y$ \quad $Y$ Summary: $\texttt{[M]}$ \\
 & & Summary: $\texttt{[M]}$ $Y$ \quad $Y$ Report: $\texttt{[M]}$ \quad Report: $\texttt{[M]}$ $Y$ \\
 \cmidrule{2-3}
 & \multirow{3}*{$Y$ It was about $\texttt{[M]}$}. & It was about $\texttt{[M]}$. $Y$ \qquad $Y$ It was around $\texttt{[M]}$. \qquad It was around $\texttt{[M]}$. $Y$ \\
 & & $Y$ It was related to $\texttt{[M]}$. \quad It was related to $\texttt{[M]}$. $Y$ \quad $Y$ It was towards $\texttt{[M]}$. \\
 & & It was towards $\texttt{[M]}$. $Y$ \\
 \cmidrule{2-3}
 & \multirow{3}*{$Y$ It was a piece of $\texttt{[M]}$ news.} & It was a piece of $\texttt{[M]}$ news. $Y$ \quad $Y$ It was a $\texttt{[M]}$ article. \quad It was a $\texttt{[M]}$ article. $Y$ \\
 &    & $Y$ It was a $\texttt{[M]}$ summary. \qquad It was a $\texttt{[M]}$ summary. $Y$ \quad $Y$ It was a $\texttt{[M]}$ report.  \\
 &    &  It was a $\texttt{[M]}$ report. $Y$ \\
 \cmidrule{2-3}
 &  \multirow{3}*{$Y$ What $\texttt{[M]}$ news!}  & What $\texttt{[M]}$ news! $Y$ \qquad $Y$ What a $\texttt{[M]}$ article! \qquad What a $\texttt{[M]}$ article! $Y$ \\
 & & $Y$ What a $\texttt{[M]}$ summary! \quad What a $\texttt{[M]}$ summary! $Y$ \quad $Y$ What a $\texttt{[M]}$ report! \\
 & &  What a $\texttt{[M]}$ report! $Y$ \\
 \midrule
 \multirow{2}*{$g(I)$} & \multicolumn{2}{l}{Verbalizer} \\
\cmidrule{2-3}
 & \multicolumn{2}{l}{$v(\textrm{Computers}, \textrm{Politics}, \textrm{Religion}, \textrm{Science})=\{(\textrm{computers}, \textrm{politics}, \textrm{religion}, \textrm{science})\}$} \\
\bottomrule
\end{tabular}}
\caption{Prompts and verbalizers used in the evaluation of attribute relevance, where $I=(X,a,Y)$ indicates the prefix, the attribute label, and the generated text, respectively.}
\label{tab:promptverb}
\end{table*}

\end{document}